\pdfoutput=1

\documentclass[11pt]{article}

\usepackage[preprint]{acl}

\usepackage{times}
\usepackage{latexsym}
\usepackage{amsmath}
\usepackage[T1]{fontenc}

\usepackage[utf8]{inputenc}

\usepackage{graphicx}
\usepackage{subfigure}
\usepackage{tabu}                   
\usepackage{multirow}                
\usepackage{multicol}                
\usepackage{multirow}                
\usepackage{float}                    
\usepackage{makecell}                 
\usepackage{booktabs}                
\usepackage{arydshln} 
\usepackage{algorithm}
\usepackage{algpseudocode}
\usepackage{amsmath} 
\usepackage{amssymb}

\usepackage{microtype}

\usepackage{inconsolata}

\usepackage{graphicx}

\title{Let's Focus on Neuron: \\ Neuron-Level Supervised Fine-tuning for Large Language Model}

\author{Haoyun Xu$^{\clubsuit\spadesuit}\thanks{Work was done while interning at Tiger Research.}\footnotemark[2]~~
$Runzhe Zhan$^{\clubsuit}$\thanks{Equal Contribution.}~~
        Derek F. Wong$^{\clubsuit}$\thanks{Corresponding Author.}~~
        \textbf{Lidia S. Chao}$^{\clubsuit}$\\
  $^{\clubsuit}$NLP$^2$CT Lab, Department of Computer and Information Science, 
  University of Macau\\
  $^{\spadesuit}$Tiger Research, Shanghai \\
  \texttt{nlp2ct.\{haoyun, runzhe\}@gmail.com, \{derekfw, lidiasc\}@um.edu.mo} \\
}

\begin{document}
\maketitle
\begin{abstract}
Large Language Models (LLMs) are composed of neurons that exhibit various behaviors and roles, which become increasingly diversified as models scale. Recent studies have revealed that not all neurons are active across different datasets, and this sparsity correlates positively with the task-specific ability, leading to advancements in model pruning and training efficiency. Traditional fine-tuning methods engage all parameters of LLMs, which is computationally expensive and may not be necessary. In contrast, Parameter-Efficient Fine-Tuning (PEFT) approaches aim to minimize the number of trainable parameters, yet they still operate at a relatively macro scale (e.g., layer-level). We introduce Neuron-Level Fine-Tuning (NeFT), a novel approach that refines the granularity of parameter training down to the individual neuron, enabling more precise and computationally efficient model updates. The experimental results show that NeFT not only exceeded the performance of full-parameter fine-tuning and PEFT but also provided insights into the analysis of neurons.

\end{abstract}

\section{Introduction}
Neurons, as fundamental components of Large Language Models (LLMs), fulfill diverse roles across model regions. As language models scale, the neurons display varying phenomena. Increase in model size enhances the the capability to generalize from text to basic and even unseen concepts \citep{DBLP:conf/iclr/PatelP22}. The internal concepts may be distributed across numerous neurons, and a significant proportion of neurons can become inactive, never triggering across diverse datasets \citep{DBLP:conf/emnlp/DurraniSDB20, DBLP:journals/corr/abs-2309-04827}. This sparsity has been substantiated by recent studies utilizing it to prune LLMs or enhance the efficiency of LLM inference, suggesting that not all neurons need to be active, a trait that becomes more pronounced in larger models \citep{DBLP:journals/corr/abs-2305-11627, DBLP:conf/icml/FrantarA23, DBLP:conf/icml/LiuWDZY0S0TRC23, DBLP:journals/corr/abs-2310-06927, DBLP:journals/corr/abs-2312-12456}. 

LLMs commonly adapt to specific tasks through full-parameter supervised fine-tuning (SFT). Parameter-efficient fine-tuning (PEFT), which operates on a layer-level modular parameter selection basis \cite{DBLP:conf/icml/HoulsbyGJMLGAG19, DBLP:conf/acl/LiL20, DBLP:conf/emnlp/LesterAC21, DBLP:conf/iclr/HuSWALWWC22}, seeks to reduce the trainable parameters during model training. Building on insights from model interpretability research, we propose that the granularity of parameter training can be refined to the neuron level. Consequently, we present a Neuron-Level Fine-Tuning (NeFT) approach, designed to improve model performance by selectively updating neurons identified as sensitive.

In an initial experiment, we identified sensitive neurons for the NLI task on Llama-2-7b-chat \citep{DBLP:journals/corr/abs-2307-09288} and used a trained probe to select a subset of neurons. The preliminary experiment suggest that training on these neurons outperform the full-parameter fine-tuning. To identify sensitive neurons for more complex tasks like translation and summarization, we devised a method that evaluates neuron similarity pre- and post-SFT, treating those with low similarity as sensitive. 
The NeFT surpasses the full-parameter fine-tuning model in task performance. To understand why it works, we further categorized neurons into three types, namely \textit{strongly affected}, \textit{suppressed}, and \textit{indirectly affected} neurons, and introduced ``rank difference'' to assess neuron utilization. 
Our findings indicate that: 1) Neurons exhibit varying degrees of sensitivity during the SFT process. 2) Neurons strongly affected by SFT elicit significant alterations in parameter utilization patterns. 3) Neurons that are important for one task tend to be relevant for others. This consistency implies that neurons identified in one context may be beneficial for transfer learning in similar datasets.

\section{Background}
\paragraph{Parameter Efficient Fine-tuning}
PEFT technique aims to enhance the performance of pre-trained models on tasks while minimizing the number of trainable parameters and computational complexity. This approach is particularly beneficial for reducing the training costs associated with large pre-trained models.

PEFT can be a empirical choice of specific layers or modules, and some studies have shown that training only a single layer can sometimes outperform full-parameter fine-tuning. For instance, \citet{DBLP:journals/corr/abs-2311-09071} found that for translation tasks, embedding fine-tuning is effective than full-parameter fine-tuning except low-resource settings. However, the empirical choice and layer-wise searching is time-consuming while recent advances focus on leveraging external module to update all the layer-level parameters. 
\paragraph{Adapter} The goal of adapter \citep{DBLP:conf/icml/HoulsbyGJMLGAG19, DBLP:conf/eacl/PfeifferKRCG21, DBLP:conf/emnlp/RuckleGGBP0G21, DBLP:conf/nips/LiuTMMHBR22} is to insert a small number of parameters into the model, and then train only these parameters when fine-tuning a downstream task, leaving the original parameters of the pre-trained model unchanged. This makes the trainable parameters more efficient and ensures that the original knowledge is not forgotten.
\paragraph{LoRA} Low-Rank Adaptation (LoRA; \citealt{DBLP:conf/iclr/HuSWALWWC22}) has emerged as one of the most prevalent methods in both academic research and industry applications. LoRA's principal concept involves decomposing a large weight matrix into two low-rank matrices, significantly reducing the number of trainable parameters. The effectiveness of LoRA depends on the chosen rank and the specific structures to which it is applied. Although the two low-rank matrices introduced by LoRA add to the model's architecture, they do not introduce additional computational costs during inference as they function concurrently with the original structures. For the sake of practical implementation, LoRA generally applied to affecting the computation of linear or multi-head attention mechanisms. 

\paragraph{Sparse Fine-Tuning} Sparse fine-tuning distinguishes itself from methods that add external modules like adapters or LoRA by introducing an initial step to pinpoint critical parameters.
This process leverages various metrics, such as Fisher information \citep{DBLP:conf/nips/SungNR21} or ${L_0}$ regularization \citep{DBLP:conf/acl/GuoRK20}, to determine which parameters are essential. These identified parameters are then specifically targeted in the subsequent training phase.
Lottery Ticket Sparse Fine-Tuning, on the other hand, targets parameters exhibiting the most significant changes during an initial epochs of full-parameter fine-tuning \citep{DBLP:conf/acl/AnsellPKV22}. 
To date, sparse training techniques, have not been thoroughly investigated within the context of LLMs \citep{DBLP:journals/corr/abs-2401-16405}, nor have they been examined from a neuron-level consideration.

\section{Preliminary Experiment}
Previous interpretability analyses within feedforward networks (FFNs) have revealed notable phenomena at the neuron level. Research by \citet{DBLP:journals/corr/abs-2310-02207} has demonstrated that certain neurons exhibit sensitivity to specific entities and can encapsulate world knowledge. 
Motivated by this insight, we begin our investigation to select certain neurons using the similar probing approach and examine their effectiveness during SFT.

\subsection{Select Neurons}
Given that NLI classification task is well-suited for probing, we choose it as our testbed of preliminary experiment to discover neurons related to the NLI task. 
We first processed the XNLI dataset\footnote{\url{https://github.com/facebookresearch/XNLI}} through the Llama-2-7b-chat model \cite{DBLP:journals/corr/abs-2307-09288} to obtain the hidden states for each layer. For each sentence in the dataset, we averaged the hidden states across all tokens, resulting in a set of hidden states for each layer. We paired each hidden state data with the corresponding target labels, i.e., Entailment, Neutral, and Contradiction, to train a Ridge classifier as a probe.
Next, we use the trained probe to identify the neurons most sensitive to XNLI tasks. 
Specifically, we selected 100,000 neurons as \textit{sensitive neurons} according to the cosine similarity between the neurons and the hidden states of probe decision space. 

\begin{figure*}[h!]
    \centering
    \includegraphics[width=0.95\linewidth]{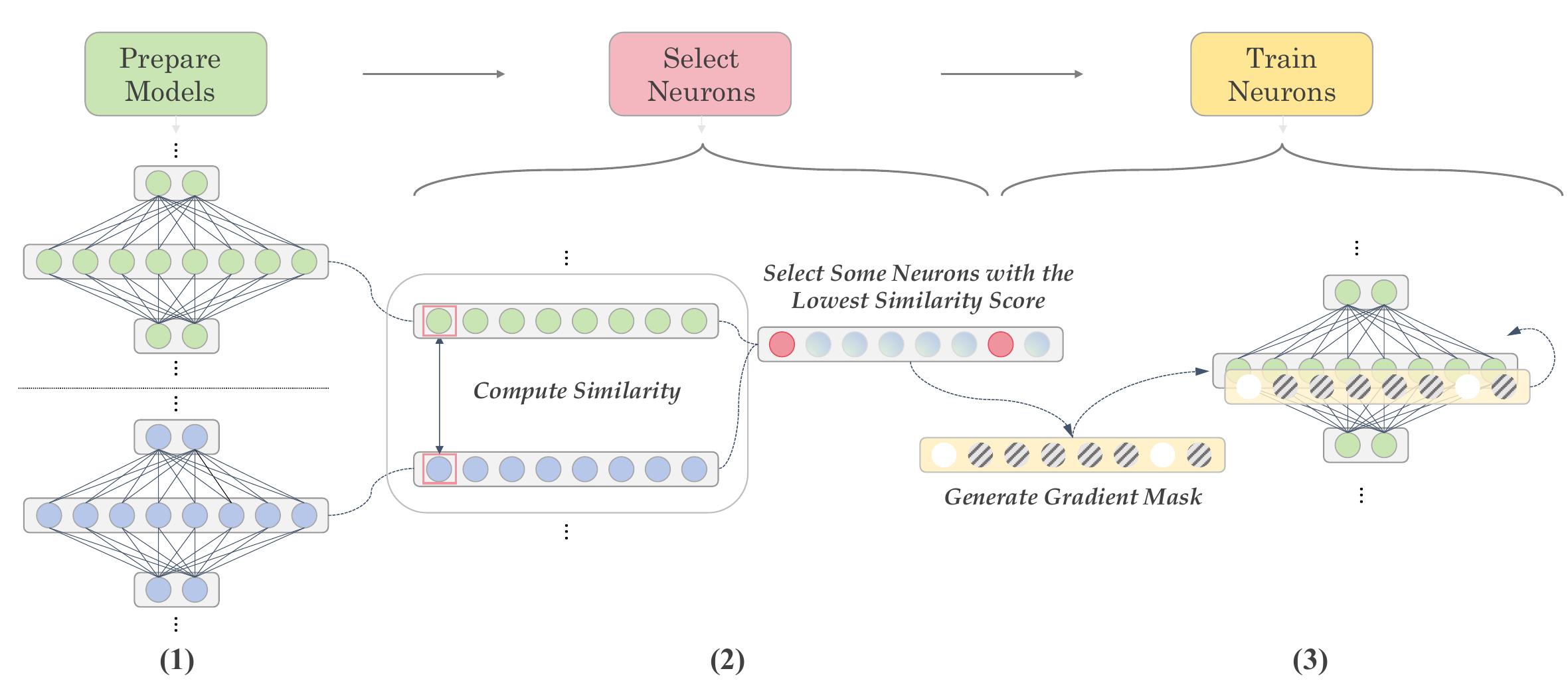}
    \caption{This diagram shows the whole process of our proposed Neuron-Level Fine-Tuning method. (1) Prepare two models, one is the original model (${{\mathbf{M}_\mathrm{Org}}}$) and the other is the model (${{\mathbf{M}_\mathrm{FT}}}$) trained with full-parameter fine-tuning. (2) Calculate the cosine similarity for each pair of neurons in the corresponding positions of ${{\mathbf{M}_\mathrm{Org}}}$ and ${{\mathbf{M}_\mathrm{FT}}}$ and select the $x\%$ neurons with the lowest score and refer to these neurons as sensitive neurons. (3) Mask the gradients of non-sensitive neurons during SFT training to ensure that only the selected neurons are updated.}
    \label{fig1}
\end{figure*}
\subsection{Experimental Results}

Upon isolating the sensitive neurons, we implement a gradient masking technique on the non-sensitive neurons throughout SFT process. 
This approach guarantees that updates are confined exclusively to the previously identified sensitive neurons.
Notably, our NeFT strategy does not require the incorporation of additional architectural elements, resulting that a mere 6\% of the total parameters were modified during training. All remaining training parameters are maintained in alignment with the full-parameter fine-tuning.

The results presented in Table \ref{table1} prove that NeFT is possible to outperform full-parameter fine-tuning by training only on a small number of sensitive neurons. 
However, this experiment on the NLI task may not fully exemplify the generality of NeFT.
In the subsequent sections, we will introduce a methodology designed to identify and train sensitive neurons in more intricate contexts, thereby extending the applicability of NeFT across a broader spectrum of tasks.

\begin{table}[h!]
\centering
\begin{tabular}{lccc}
\toprule
 & \textbf{English} & \textbf{French} & \textbf{German} \\
  \midrule
 \textbf{FT-full} & 82.8 & 70.7 & 67.0 \\
 \textbf{NeFT} & \textbf{84.9} & \textbf{76.9} & \textbf{80.2} \\
 \bottomrule
\end{tabular}
\caption{Preliminary experiment of NeFT effectiveness was conducted on XNLI tasks. NeFT demonstrated superior performance compared to full-parameter training.}
\label{table1}
\end{table}

\section{Methodology}
Current probing techniques fall short in addressing complex tasks. To remedy this, we propose a flexible approach as shown in Figure \ref{fig1} and Algorithm \ref{alg}. to facilitate the identification of sensitive neurons that are most influential to model performance improvements.

\subsection{Prepare Models}
In fact, training a model entails the discernment and engagement of sensitive neurons. By evaluating neurons that deviate most from their initial state, we can ascertain which ones the model has prioritized during its training, thereby revealing their sensitivity to the task at hand. Building on this premise, we begin by preparing an initial model alongside its fine-tuned counterpart for the designated task. 
We employ the SFT dataset $\mathcal{D}$ to train initial model ${\mathbf{M}_\mathrm{Org}}$ with all parameters, resulting in the fine-tuned model ${{\mathbf{M}_\mathrm{FT}}}$. In practice, for the sake of efficiency, the fine-tuned model also can be derived within a limited number of training steps. Our analysis in subsequent sections suggests that this abbreviated training regimen yields a model whose performance is not significantly different from one that has been trained to convergence.

\subsection{Select Neurons by Model Itself}
We define an individual neuron as a distinct entry within the weight matrix. For example, in a linear layer of an MLP with a weight matrix of dimension $\mathbb{R}^{m \times n}$, a single neuron corresponds to a row in this matrix, which is dimensionally represented as $\mathbb{R}^{1 \times n}$.
We then calculate the cosine similarity for each corresponding pair of neurons between the original model and the fine-tuned model. 
A neuron at a given position $(i,n)$ within the models is denoted by $\mathbf{W}_\mathrm{Org}^{(i,n)}$ and $\mathbf{W}_\mathrm{FT}^{(i,n)}$, where ${i}$ represents the layer index and ${n}$ indicates the ${n}$-th neuron within that layer.
Following the formula in Equation \ref{cos_sim}, we calculate the cosine similarity score $\{{{S}^{(i,n)}\}}$:

\begin{equation}
S(w_\mathrm{Org}, w_\mathrm{FT}) = \frac{w_\mathrm{Org} \cdot w_\mathrm{FT}}{\lvert \lvert w_\mathrm{Org} \rvert \rvert \times \lvert \lvert w_\mathrm{FT} \rvert \rvert}
\label{cos_sim}
\end{equation}
\noindent where  ${w_\mathrm{Org}}$ ${\in}$ $\mathbf{W}_\mathrm{Org}^{(i,n)}$, ${w_\mathrm{FT}}$ ${\in}$ $\mathbf{W}_\mathrm{FT}^{(i,n)}$.
The $k$ neurons with the lowest cosine similarity scores are then selected for further training process.

\subsection{Neuron-Level Fine-Tuning}
After calculating the similarity scores for each neuron pair, we rank the neurons according to their similarity scores to pinpoint the ones to which the model allocates the most attention during training, characterized by their lower similarity scores. We record the location of these neurons, represented by the positional information $(i,n)$.
To train solely the identified neurons, we modify the gradient $\mathbf{g}(\mathbf{W}_\mathrm{Org})$. Prior to each update during the training process, we refer to the previously saved positional information $(i,n)$ to decide if the gradient of a specific neuron should be retained or set to zero. This gradient masking policy ensures that only the specific neurons are updated.

\begin{algorithm}
\caption{Neuron-Level Fine-Tuning (NeFT)}
\begin{algorithmic}[1]
\Require SFT Dataset $\mathcal{D}$, model ${\mathbf{M}_\mathrm{Org}}$. 
\State Train full parameters of ${\mathbf{M}_\mathrm{Org}}$ on dataset $\mathcal{D}$ within limited steps $K$ and obtain ${\mathbf{M}_\mathrm{FT}}$.
\State Calculate similarity scores $\{{{S}^{(i,n)}\}}$ by Eq.\ref{cos_sim}.
\State Select $x\%$ of the neurons $\{w^{(i,n)}\} \leftarrow \arg\min_{S^{(i,n)}} \mathrm{Select}({\mathbf{W}_\mathrm{Org}}, \{i\}, \{n\})$.
\For{$1,\dots,\mathrm{Epochs}$}
    \For{$1,\dots,\mathrm{Batches}$}
    \State Calculate gradient $\mathbf{g}$ of each batch.
    \State $\mathbf{g}(\{\mathbf{W}_\mathrm{Org} \setminus \{w^{(i,n)}\}\}) \leftarrow 0$
    \State Back propagate the gradient $\mathbf{g}$.
    \EndFor
\EndFor
\end{algorithmic}
\label{alg}
\end{algorithm}

\begin{table*}[h!]
\centering
\scalebox{0.88}{
\begin{tabular}{lccccccccccc}
\toprule
 & & \multicolumn{2}{c}{\textbf{En${\rightarrow}$Zh} } & \multicolumn{2}{c}{\textbf{En${\rightarrow}$Fr}} & \multicolumn{2}{c}{\textbf{Fr${\rightarrow}$Zh}} & \multicolumn{1}{c}{\textbf{Hi${\rightarrow}$Zh}} & \multicolumn{1}{c}{\textbf{Hi${\rightarrow}$Fr}} & \multicolumn{1}{c}{\textbf{En${\rightarrow}$Mi}} & \multicolumn{1}{c}{\textbf{En${\rightarrow}$Bs}} \\
\cmidrule(lr){3-4}\cmidrule(lr){5-6}\cmidrule(lr){7-8}\cmidrule(lr){9-9}\cmidrule(lr){10-10}\cmidrule(lr){11-11}\cmidrule(lr){12-12}

\textbf{Method} & \textbf{\%$_\text{Para.}$} &\textbf{20k} & \textbf{100k} & \textbf{20k} & \textbf{100k} &  \textbf{20k} & \textbf{100k} & \textbf{4.5k} & \textbf{4.5k} & \textbf{10k} & \textbf{10k}\\
\midrule

\textbf{FT-full} & 100 &22.22 & 26.42 & 26.65 & 31.00 &  18.20 & 22.22 &  6.21 &  6.37 &  14.50 &  7.35 \\

\textbf{FT-embed} & 2 &23.51 & 25.40 &  29.71 & 30.85 &  19.33 & 20.82 &  6.91 &  6.78 & 2.37 & 4.33 \\

\textbf{FT-\{in | out\}} & 41 & 24.35 & 27.35 & 28.70 & 33.65 & 20.91 & 22.78 & 9.03 & 7.55 & \textbf{15.11} & 7.82 \\

\textbf{LoRA$_{r=256}$} & 9 &27.15 & 26.22 & 33.14 & 33.52 & 22.86 & 22.25 & 8.64 & 5.66 & 4.55 & 8.22 \\

\textbf{NeFT$_{9\%}$} & 9 &\textbf{28.70} & \textbf{30.48} & \textbf{34.86} & \textbf{36.71} & \textbf{24.08} & \textbf{25.62} & \textbf{10.69} & \textbf{10.23} & 13.90 & \textbf{10.04} \\
\bottomrule
\end{tabular}
}
\caption{
Performance comparison of each method on the translation task.
For LoRA and NeFT, we report the performance when tuning 9\% of model parameters.
In the majority of cases, NeFT outperforms the baseline methods. More detailed results are available in Appendix Table \ref{table10}.
\label{table2}
}
\end{table*}

\begin{table*}[h!]
\centering
\scalebox{0.9}{
\begin{tabular}{lcccccccccc}
\toprule
& &\multicolumn{3}{c}{\textbf{En${\rightarrow}$Zh}} & \multicolumn{3}{c}{\textbf{Fr${\rightarrow}$Zh}} & \multicolumn{3}{c}{\textbf{Hi${\rightarrow}$Zh}} \\
\cmidrule(lr){3-5}\cmidrule(lr){6-8}\cmidrule(lr){9-11}
 & & \multicolumn{3}{c}{\textbf{2.3k}} &  \multicolumn{3}{c}{\textbf{0.1k}} &  \multicolumn{3}{c}{\textbf{0.1k}} \\
\textbf{Method} & \textbf{\%$_\text{Para.}$} & R1 & R2 & RL & R1 & R2 & RL & R1 & R2 & RL\\
\midrule
\textbf{FT-full} & 100 &21.88 & 17.39 & 20.71 & 30.24 & 19.79 & 25.36 & 10.16 & 4.35 & 9.42 \\
\textbf{FT-embed} & 2 & 1.35 & 0.67 & 1.09 & 10.48 & 6.39 & 8.32 & 0.73 & 0.19 & 0.67\\
\textbf{FT-\{in | out\}} & 41 & 23.13 & 18.16 & 22.07 & 32.45 & 26.59 & 29.24 & 15.51 & 8.21 & 13.60 \\
\textbf{LoRA$_{r=256}$} & 9 & 19.94 & 15.73 & 18.74 & 4.48 & 2.72 & 4.48 & 2.46 & 0.45 & 2.16 \\
\textbf{NeFT$_{9\%}$} & 9 & 24.31 & 19.27 & 23.29 & \textbf{33.57} & \textbf{26.67} & \textbf{31.46} & 15.23 & \textbf{9.70} & 14.21 \\
\textbf{NeFT$_{12\%}$} & 12 & 23.38 & 18.71 & 22.54 & 29.96 & 22.26 & 25.33 & 13.56 & 8.12 & 12.58 \\
\textbf{NeFT}${_{9\%\mathrm{Union}}}$ & 12 & \textbf{26.05} & \textbf{21.06} & \textbf{25.07} & 30.85 & 22.86 & 26.60 & \textbf{15.78} & 9.32 & \textbf{14.36} \\
\bottomrule
\end{tabular}
}
\caption{
Performance comparison of each method on cross-lingual summarization task. 
For comparing LoRA with NeFT, we report the performance when tuning 9\% of model parameters.
The results indicate that NeFT remains effective even in low-resource settings.
} 
\label{table3}
\end{table*}

\section{Experiments} 
\subsection{Experimental Settings}
\paragraph{Data}
Our primary experiments are centered on machine translation and cross-lingual text summarization tasks. 
For the machine translation task, we sourced our training and development datasets from the News Commentary \citep{DBLP:conf/lrec/Tiedemann12} and Lego-MT \citep{DBLP:conf/acl/YuanLZKLQX23}. 
The test set was collated from Flores-101 \citep{DBLP:journals/tacl/GoyalGCCWJKRGF22} and Lego-MT, ensuring a diverse range of linguistic challenges. 
Regarding the cross-lingual text summarization task, our dataset was taken from CrossSum \citep{DBLP:conf/acl/BhattacharjeeHA23}. Due to computational resource limitations, we restricted our training to sentences comprising fewer than 1024 tokens, thereby targeting on a low-resource setting.

\paragraph{Training Setup}
We conducted our experiments on the Llama-2-7b-chat model.
We fine-tuned all models with a batch size of $6$ for translation tasks and a batch size of $3$ for summarization tasks until they reached convergence. The optimal checkpoints were determined based on the lowest validation loss observed on the development datasets. Typically, the checkpoint corresponding to the lowest evaluation loss emerged from the first epoch. However, in the summarization task, particularly when dealing with smaller datasets, this convergence point might occur during the second epoch.
For training with LoRA, convergence was typically attained within 2 to 3 epochs. In the case of NeFT, when a certain percentage ${x\%}$ of the model parameters were trained, we denoted this as NeFT${_{x\%}}$. Similarly, for LoRA configurations, if the LoRA rank was set to \textit{r}, we represented this as LoRA$_{r}$.

\paragraph{Baselines}
Our method was evaluated against four baselines:
\begin{itemize}
    \item \textbf{FT-full}: This strategy fine-tunes all parameters of the LLM.
    \item \textbf{FT-\{in | out\}}: This approach fine-tunes the weights of input and output projection layers of all the MLPs.
    \item \textbf{FT-embed} \cite{DBLP:journals/corr/abs-2311-09071}: This technique focuses solely on fine-tuning the embedding layer.
    \item \textbf{LoRA} \cite{DBLP:conf/iclr/HuSWALWWC22}: We apply it to fine-tune all linear structures within each layer (excluding the head layer of language model).
\end{itemize}
To maintain the integrity of the comparison, we adjusted the number of trainable parameters of NeFT to closely align with those employed in the LoRA configurations.

\paragraph{Evaluation}
We employ the ordinary BLEU score \cite{DBLP:conf/acl/PapineniRWZ02} to assess the quality of machine translations, which measures the similarity of n-grams between the machine-generated text and the reference translation. Higher BLEU scores indicate better translation quality.
In the context of text summarization, we utilize the ROUGE metric \citep{lin-2004-rouge} to evaluate the extent to which the machine-generated summary encapsulates the core points of the reference summary. ROUGE assesses the recall rate of the generated summaries by calculating the overlap of $N$-grams, which are represented as ``R($N$)'' in the reported results, where $N$ corresponds to the length of the $N$-gram.

\subsection{Main Results}
\paragraph{Performance and Generalization}
In Table \ref{table2} and \ref{table3}, we present PEFT performance for each method. It is evident that NeFT outperforms the other methods in the majority of cases, regardless of task and language pair. Notably, in machine translation experiments across various data scales, we fine-tune the same subset of neurons.

In addition to its superior performance on targeted language pairs, we also find that NeFT have a potential to improve cross-lingual generalization. 
For summarization task, we merge an equal number of neurons identified from the translation task of a specific language pair with those from the summarization task. For instance, neurons from the NeFT${_{6\%}}$ configuration in the summarization task were combined with those from the NeFT${_{6\%}}$ setting in the translation task. This amalgamated configuration is referred to as NeFT${_{6\%\mathrm{Union}}}$. 
Our observations indicate that the potential for further improving cross-lingual summarization with neurons identified in translation task is more pronounced when dealing with high-resource language pairs like English-to-Chinese translation.

\begin{figure}
    \centering
    \includegraphics[width=\linewidth]{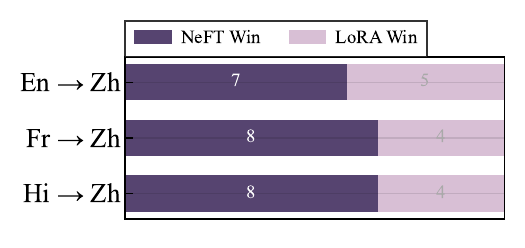}
    \caption{Comparison of NeFT and LoRA across different trainable parameter settings. NeFT consistently utilizes fewer parameters than LoRA at each level. The details are presented in Appendix Table \ref{table10}.}
    \label{fig_compare}
\end{figure}

\paragraph{Compare with LoRA} 
We conducted a comprehensive comparison with LoRA as shown in Figure \ref{fig_compare}.
We varied LoRA rank between 8 and 256 in order to maintain a comparable number of trainable parameters with NeFT neurons.
Our findings reveal that NeFT outperforms LoRA in terms of translation quality, particularly in the X-to-Chinese translation direction. For detailed comparisons and scores, please refer to Appendix Table \ref{table10}.

\section{Analysis}
\subsection{Utilization of Neurons}
\paragraph{Metric} To explore the divergence of neuron utilization, we compared models trained under various NeFT configurations by analyzing neuron utilization. 
During inference process, we retained the hidden states for each layer and calculated the Pearson correlation coefficient for each neuron, recording the highest value per neuron. We then ordered the neurons by their maximum Pearson score in descending order, denoting the ranking of neurons as $\mathbf{Rank}$.
To discern the contrast in neuron utilization between models, we compute the rank disparity for each neuron across the two models, represented as ${\Delta\mathbf{Rank}}$. 
We then determined the mean of the absolute values of these rank differences, ${\mathrm{Avg}(\Delta\mathbf{Rank})}$, to quantify the overall divergence in utilization of the neurons.

\begin{figure}[h!]
    \centering
    \includegraphics[width=\linewidth]{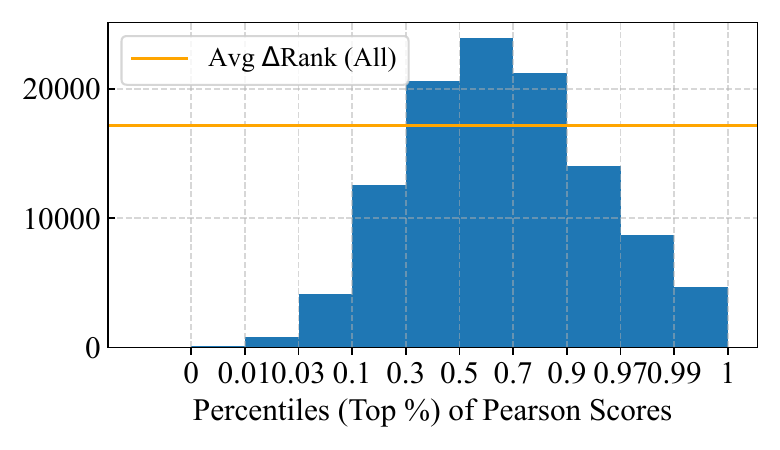}
    \caption{Average rank differences between NeFT${_{6\%}}$ and NeFT${_{3\%}}$ were calculated for neurons. The ranks were sorted based on their pairwise Pearson scores in descending order.}
    \label{fig2}
\end{figure}

\paragraph{Shifts of Neuron Utilization} 
To analyze how utilization of neurons shift after fine-tuning, we categorize the neurons into different buckets according their Pearson correlation scores and calculate the average rank differences of each bucket between NeFT${{_6\%}}$ and NeFT${{_3\%}}$, denoted as ${\mathrm{Avg}(\Delta\mathbf{Rank})}$. The bucketing strategy relies on the top percentiles of the overall Pearson scores.
Figure \ref{fig2} demonstrates that the shift in utilization for neurons within the top 0\% to 0.03\% range is negligible, whereas neurons with intermediate correlation scores exhibit the most significant sensitivity to fine-tuning.
These findings suggest that our selection strategy of sensitive neurons is reasonable. 
Furthermore, we also found that this observation is consistent across various training settings for other NeFT models.

\subsection{Effects of Neuron Selection Settings}
\begin{figure}
    \centering
    \includegraphics[width=\linewidth]{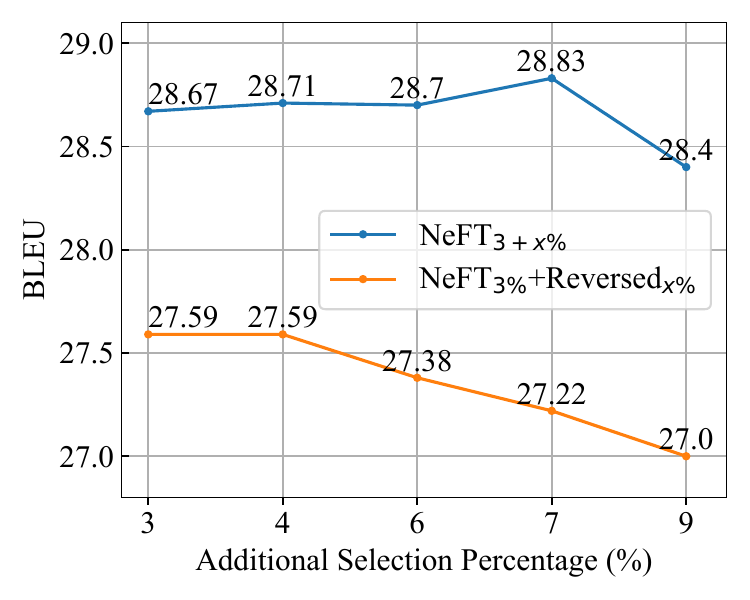}
    \caption{BLEU scores of models trained with different NeFT settings. By using NeFT${_{3\%}}$ as a base setting, neurons have high similarity scores and those with low similarity scores were separately incorporated and trained using 20k English-Chinese translation data.}
    \label{fig_adding_neurons}
\end{figure}

\paragraph{Performance Comparison} To investigate the impact of neuron selection based on similarity score on NeFT training, we incorporated neurons with both high and low cosine similarity scores into the NeFT${_{3\%}}$ model and examined the respective performance trends. The study used a dataset comprising 20,000 English-to-Chinese translation training instances.
Adding neurons with high similarity scores were labeled Reversed${_{x\%}}$ for clarity. Hence, the notation ``NeFT${_{3\%}}$+Reversed${_{x\%}}$'' represents a hybrid selection strategy by combining neurons from NeFT${_{3\%}}$ and Reversed${_{x\%}}$. According to the results presented in Figure \ref{fig_adding_neurons}, the original NeFT neuron selection strategy consistently outperforms the contrasting strategy. Moreover, it also shows a decline in model performance as the proportion of high-similarity neurons increases.

\begin{figure}[h!]
    \centering
    \includegraphics[width=1\linewidth]{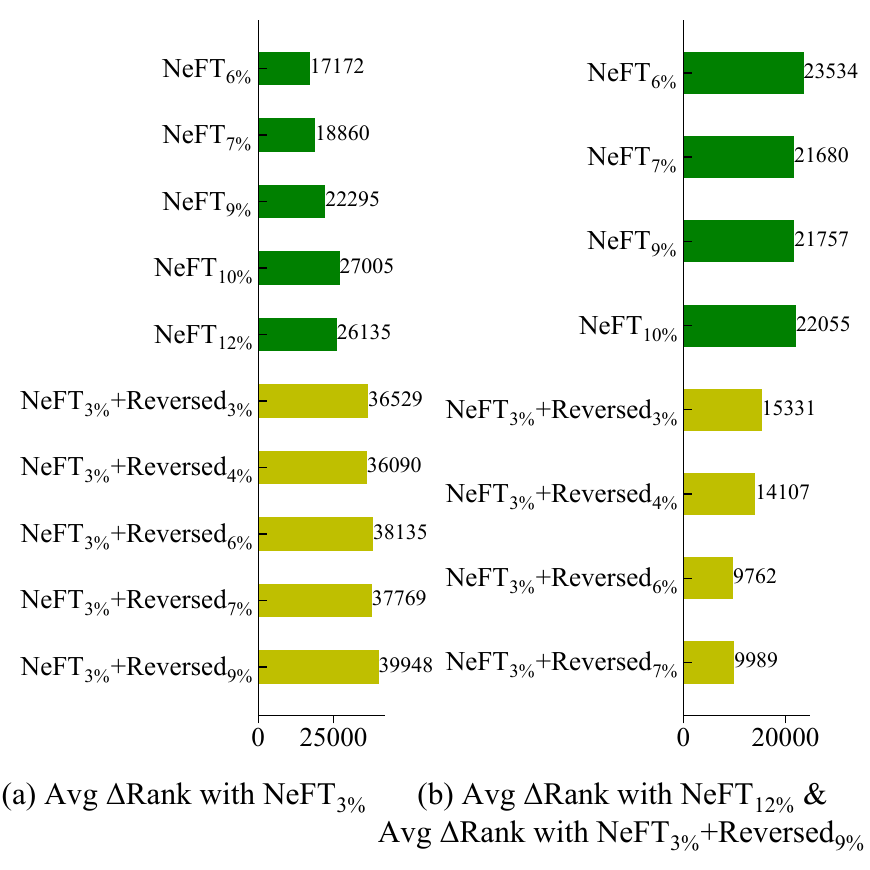}
    \caption{Rank difference
    ${\mathrm{Avg}(\Delta\mathbf{Rank}})$ is calculated in order to assess the shifts in the utilization of neurons. Overall, the neuron utilization of original neuron selection strategy NeFT${_{x\%}}$ is more stable than that of contrasting selection strategy NeFT${_{3\%}}$+Reversed${_{x\%}}$.
    }
    \label{fig3}
\end{figure}

\paragraph{Dynamics of Neuron Utilization}
Figure \ref{fig3} provides insights into how neuron utilization is influenced by different NeFT training configurations.
In Figure \ref{fig3} (a), the comparison with the NeFT${_3\%}$ model which employs the original selection strategy, shows that neuron utilization discrepancies becomes more pronounced when the model is trained with high-similarity neurons. This observation implies that the contrasting strategy, NeFT${{_3\%}}$+Reversed${_x\%}$, tends to induce greater volatility within the neuron utilization patterns.
Figure \ref{fig3}(b) analyzes the differences in utilization between models with a larger proportion of trainable parameters and those with a smaller proportion, while employing the same fine-tuning strategy. This comparison is made between both the original selection strategies (e.g., NeFT${{_{12}\%}}$ vs. NeFT${{_x\%}}$) and the contrasting selection strategies (e.g., NeFT${{_3\%}}$+Reversed${{_9\%}}$ vs. NeFT${{_3\%}}$+Reversed${{_x\%}}$).
From the observed data, different from Figure \ref{fig3} (a), it is apparent that models employing the contrasting neuron selection strategy are less affected by an increase in the number of trainable parameters. This suggests that the original neuron selection strategy has a higher degree of stability with respect to changes in the scale of trainable parameters.

\paragraph{Category of Neurons}
Based on the observed ranking shifts, we further classify neurons into one of three different categories: ``Strongly Affected Neurons'', ``Suppressed Neurons'', and ``Indirectly Affected Neurons''.
\begin{itemize}
    \item \textbf{Strongly Affected Neurons}: Neurons exhibiting a rank difference ${\Delta\mathbf{Rank}}$ exceeding 100,000 are categorized as \textit{Strongly Affected Neurons}, indicating significant influence from NeFT training.
    \item \textbf{Suppressed Neurons}: Within the subset of strongly affected neurons, not all exhibited upward movement in their rankings. Consequently, we classify neurons that decreased in rank as \textit{Suppressed Neurons}.
    \item \textbf{Indirectly Affected Neurons}: The training of a specific subset of neurons inherently impacts the remaining, untrained neurons during the inference process. Within the subset of strongly affected neurons, we observe that a substantial number of strongly affected neurons were not directly engaged in NeFT training. These neurons are thus designated as \textit{Indirectly Affected Neurons}.
\end{itemize}

\begin{figure}[h!]
    \centering
    \includegraphics[width=1\linewidth]{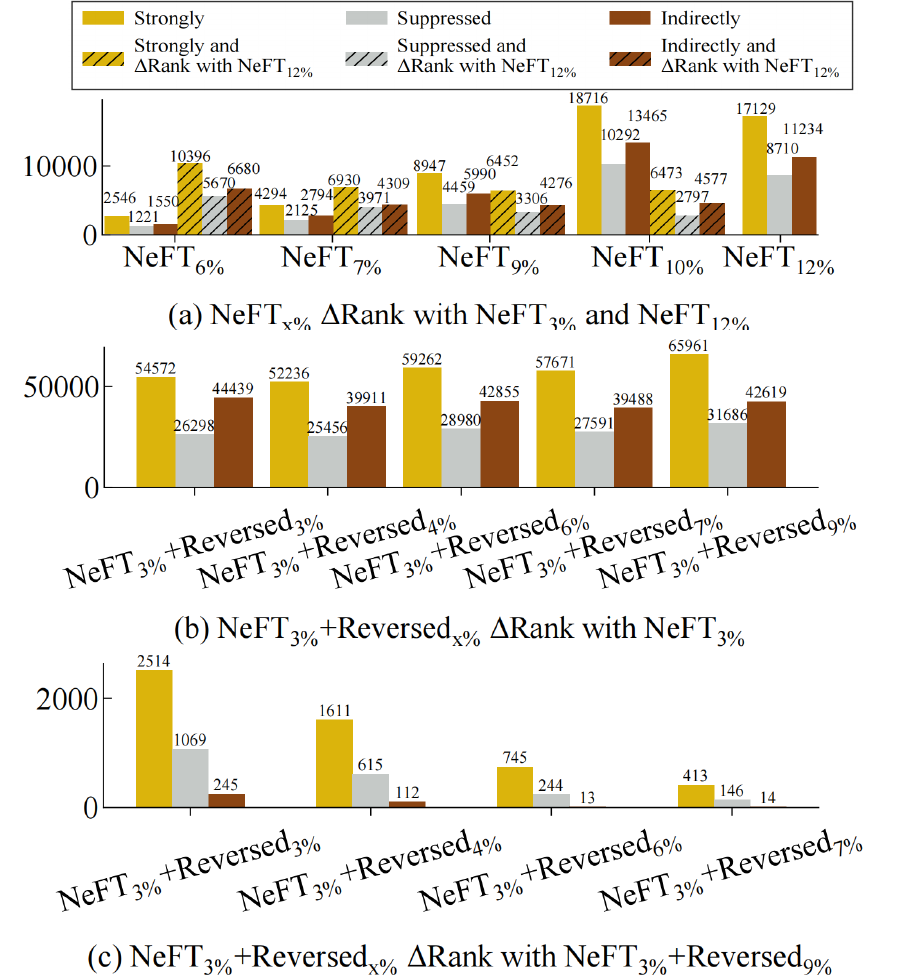}
    \caption{The distribution of three types of neurons (Strongly affected, Suppressed, and Indirectly affected) across models under various training settings.}
    \label{fig_different_neurons}
\end{figure}

\paragraph{Categorized Comparison} 
Additionally, we assessed the quantity of different neuron types within the original selection strategy NeFT${_{x\%}}$ and the contrasting strategy NeFT${{_3\%}}$+Reversed${_{x\%}}$. The categorization includes strongly affected neurons, suppressed neurons, and indirectly affected neurons, as illustrated in Figure \ref{fig_different_neurons}.
The trend indicates a higher count of strongly affected neurons in the contrasting strategy compared to the original selection strategy (Figure \ref{fig_different_neurons} (a) and Figure \ref{fig_different_neurons} (b)). In the case of suppressed neurons, there is a noticeable trend: as the number of strongly affected neurons rises, the quantity of suppressed neurons increases correspondingly, maintaining a consistent proportion.
In Figure \ref{fig_different_neurons} (c), the number of strongly affacted neuron found between two NeFT${{_3\%}}$+Reversed${_{x\%}}$ settings is much lower than that between NeFT${{_3\%}}$+Reversed${_{x\%}}$ and NeFT${{_3\%}}$. Also, the proportion of indirectly affected neurons is very low which different from Figure \ref{fig_different_neurons} (a) and Figure \ref{fig_different_neurons} (b).

Based on the above observations discussed in this section, we infer that the inclusion of neurons with high similarity score causes drastic changes in the model's neuronal utilization patterns. However, as the number of neurons increases, the extent of change becomes relatively small. Additionally, the trend observed with the inclusion of more neurons may imply that only a minimal number of \textit{strongly affected neurons} are subject to indirect influence, leading to minimal shifts in the neurons' utilization.

\subsection{Ablation Study}

\begin{table}[h!]
\centering
\scalebox{0.9}{
\begin{tabular}{lccc}
\toprule
\textbf{Original Data(20k)} & \\
\midrule
\textbf{FT-full} & 22.22 \\
\textbf{NeFT${_{9\%}}$} & \textbf{28.70} \\
\textbf{NeFT}${_{9\%\mathrm{Early}}}$ & 28.51 \\
\midrule
\textbf{Separated Data (20k)} & \\
\midrule
\textbf{FT-full} & 22.12 \\
\textbf{NeFT${_{6\%}}$} & 28.46 \\
\textbf{NeFT}${_{6\%\mathrm{Trans}}}$ & \textbf{28.57} \\
\bottomrule
\end{tabular}
}
\caption{\label{table7} BLEU scores of NeFT-tuned English-Chinese translation model under different ablation settings.
}
\end{table}

\paragraph{Neuron Selection} We also focus on assessing the impact of neurons selected by the model prior to its convergence. We selected neurons from a model that had been trained on the English-to-Chinese (En-Zh) translation dataset for only 800 steps, and these neurons are denoted as NeFT${_{x\%\mathrm{Early}}}$ and would be used to select sensitive neurons. As shown in Table \ref{table7}, the neurons identified by the unconverged model also can delivers satisfactory results, indicating that effective neuron selection does not necessarily require a fully converged model.

\paragraph{Neuron Generalization} To explore the influence of identified neurons on additional datasets within the same domain, we created an additional English-to-Chinese (En-Zh) translation training dataset. We then applied the neurons identified from the original dataset, denoted as NeFT${_{x\%\mathrm{Trans}}}$, to train models on these unseen data instances.
The results in Table \ref{table7} indicate that neurons, once identified by the model, can be effectively employed across similar datasets, maintaining consistent performance. This suggest the potential for neurons identified in one context to be extrapolated to other contexts within the same domain.

\paragraph{Overlap Analysis}
We conducted further analysis on the overlap of neurons identified under different settings as per the ablation experiments previously described. The data presented in Table \ref{table7} reveal that while the specific neurons selected across different settings are not identical, there is a considerable degree of overlap. Furthermore, the extent of this overlap escalates with an increase in the number of neurons designated for selection. Despite some variability in the specific neurons identified, the overall impact on the performance of models remains largely consistent, as evidenced by the previous results.
\begin{table}[h!]
\centering
\scalebox{0.65}{
\begin{tabular}{lcccc}
\toprule
\multicolumn{1}{l}{\textbf{Neurons Overlap}} & \multicolumn{4}{c}{\textbf{NeFT${_{x\%}}$}} \\
 & \textbf{3\%} & \textbf{6\%} &\textbf{9\%} &\textbf{12\%} \\ 
\midrule
\textbf{Original (20k) | Separated (20k)} & 63\% & 68\% & 72\% & 76\% \\
\textbf{Original (20k) | Original (100k)} & 61\% & 65\% & 68\% & 71\% \\
\textbf{Convergence (20k) | Non-convergence (20k)} & 63\% & 68\% & 72\% & 75\% \\

\bottomrule
\end{tabular}
}
\caption{\label{table8}
Overlap proportion of neurons across different data and training settings.
}
\end{table}

Taking into account the insights from the preceding sections, we can deduce that the crucial factor regarding neuron selection is not the inclusion of a substantial quantity of non-sensitive Reversed${{_x\%}}$ neurons. Regarding NeFT${{_x\%}}$ neurons, the selection criteria do not require excessive precision, approximate selections are found to be adequate.

\section{Conclusion}
In this study, we introduced Neuron-level Supervised Fine-tuning, NeFT, an innovative approach to supervised fine-tuning. NeFT focuses on identifying task-specific neurons within a model by evaluating the similarity between the neurons of the trained and the original model. By selecting and training on neurons that exhibit low similarity.
Our empirical results demonstrate that NeFT generally surpasses full-parameter fine-tuning and other fine-tuning methodologies across various settings. When compared with LoRA, NeFT consistently shows superior or comparable results at most levels of parameter counts.
Furthermore, our analyses shed light on the functioning of distinct neurons in the NeFT process and the impact of various settings. This understanding contributes to a more nuanced view of the fine-tuning process and opens avenues for further optimization and refinement of neuron-level training strategies. In the future, we will keep applying NeFT method to more  generation and reasoning tasks.

\section*{Limitation}
This work encounters certain constraints due to the capabilities of current distributed frameworks, particularly regarding gradient operations. Specifically, we are unable to apply gradient operations universally across all structural components of the model. Consequently, our experiments are confined to the ${\mathbf{up\_proj}}$ and ${\mathbf{down\_proj}}$ projections within the Llama-2-7b-chat architecture, and we have not extended our investigation to include the ${\mathbf{gate\_proj}}$ projections. Despite these restrictions, the NeFT methodology has demonstrated impressive results.
For future analyses, there are existing neuron-pruning methods that treat neurons as discrete units, as referenced by \citep{DBLP:conf/cvpr/FangMSMW23,DBLP:conf/nips/ZhuangZHZSL20,DBLP:journals/corr/HanPTD15,DBLP:journals/corr/abs-2306-11695,DBLP:journals/corr/abs-2305-11627}. These methods could potentially be leveraged in model pruning to eliminate certain neurons. This would enable a more concentrated examination of the neurons that contribute the most value to the model's performance, thereby optimizing the fine-tuning process by focusing on the most impactful neurons.

\section*{Acknowledgements}
This work was supported in part by the Science and Technology Development Fund, Macau SAR (Grant Nos. FDCT/060/2022/AFJ, FDCT/0070/2022/AMJ), National Natural Science Foundation of China (Grant No. 62261160648), Ministry of Science and Technology of China (Grant No. 2022YFE0204900), and the Multi-year Research Grant from the University of Macau (Grant No. MYRG-GRG2023-00006-FST-UMDF). This work was performed in part at SICC which is supported by SKL-IOTSC, and HPCC supported by ICTO of the University of Macau.

\bibliography{custom}
\newpage
\appendix

\section{Appendix}
\label{sec:appendix}
Table \ref{table10} presents the results of NeFT and LoRA under various training settings for translation of each language pair. It also includes a comparison between NeFT and LoRA at similar trainable parameter scales. 

\begin{table*}[h!]
\centering
\scalebox{0.73}{
\begin{tabular}{lccccccc}
\toprule
\textbf{\# Lang En${\rightarrow}$Zh (20k)} & \textbf{BLEU} & & & & \textbf{BLEU} \\
\textbf{NeFT | Params} & \textbf{En${\rightarrow}$Zh} & \textbf{Fr${\rightarrow}$Zh}& \textbf{Hi${\rightarrow}$Zh} & \textbf{LoRA | Params} & \textbf{En${\rightarrow}$Zh} & \textbf{Fr${\rightarrow}$Zh}& \textbf{Hi${\rightarrow}$Zh}\\
\midrule
& & & & \textbf{Rank 8 | 20.0M} & 27.13 & \underline{23.05} & 2.85 \\
\textbf{NeFT${_{0.4\%}}$ | 25.6M} & 27.21 & 22.74 & 2.40 & \textbf{Rank 16 | 40.0M} & \underline{27.29} & 23.04 & \textbf{\underline{2.88}} \\
\hdashline[1pt/5pt]
\textbf{NeFT${_{1.5\%}}$ | 102.4M} & \underline{28.32} & \underline{23.30} & \underline{2.17} & \textbf{Rank 64 | 159.9M} & 27.34 & 22.21 & 0.90 \\
\hdashline[1pt/5pt]
\textbf{NeFT${_{3\%}}$ | 204.8M} & \underline{28.34} & \underline{23.78} & \underline{0.99} & \textbf{Rank 128 | 319.8M} & 27.27 & 22.81 & 0.20 \\
\hdashline[1pt/5pt]
\textbf{NeFT${_{6\%}}$ | 409.6M} & 28.67 & 23.76 & \underline{0.81} & \textbf{Rank 256 | 639.6M} & 27.15 & 22.60 & 0.09 \\
\textbf{NeFT${_{9\%}}$ | 614.4M} & \textbf{\underline{28.70}} & \textbf{\underline{24.02}} & 0.46 \\
\hdashline[1pt/5pt]
\textbf{NeFT${_{12\%}}$ | 819.2M} & 28.40 & 23.80 & 1.30 \\
\midrule
\midrule

\textbf{\# Lang Fr${\rightarrow}$Zh (20k)} & \textbf{BLEU} & & & & \textbf{BLEU} \\
\textbf{NeFT | Params} & \textbf{En${\rightarrow}$Zh} & \textbf{Fr${\rightarrow}$Zh}& \textbf{Hi${\rightarrow}$Zh} & \textbf{LoRA | Params} & \textbf{En${\rightarrow}$Zh} & \textbf{Fr${\rightarrow}$Zh}& \textbf{Hi${\rightarrow}$Zh}\\
\midrule

& & & & \textbf{Rank 8 | 20.0M} & 24.19 & 22.77 & 1.18 \\
\textbf{NeFT${_{0.4\%}}$ | 25.6M} & \underline{24.49} & \underline{23.00} & \textbf{\underline{8.61}} & \textbf{Rank 16 | 40.0M} & 24.27 & 22.55 & 1.71 \\
\hdashline[1pt/5pt]
\textbf{NeFT${_{1.5\%}}$ | 102.4M} & \underline{24.99} & \underline{23.31} & \underline{7.94} & \textbf{Rank 64 | 159.9M} & 24.57 & 22.57 & 0.62 \\
\hdashline[1pt/5pt]
\textbf{NeFT${_{3\%}}$ | 204.8M} & \underline{24.95} & \underline{23.67} & \underline{8.09} & \textbf{Rank 128 | 319.8M} & 20.87 & 22.83 & 0.56 \\
\hdashline[1pt/5pt]
\textbf{NeFT${_{6\%}}$ | 409.6M} & \textbf{\underline{25.32}} & 23.65 & 6.36 & \textbf{Rank 256 | 639.6M} & 24.51 & 22.86 & 0.57 \\
\textbf{NeFT${_{9\%}}$ | 614.4M} & 24.74 & \textbf{\underline{24.08}} & \underline{6.96} \\
\hdashline[1pt/5pt]
\textbf{NeFT${_{12\%}}$ | 819.2M} & 23.90 & 23.97 & 3.34 \\
\midrule
\midrule

\textbf{\# Lang Hi${\rightarrow}$Zh (4.5k)} & \textbf{BLEU} & & & & \textbf{BLEU} \\
\textbf{NeFT | Params} & \textbf{En${\rightarrow}$Zh} & \textbf{Fr${\rightarrow}$Zh}& \textbf{Hi${\rightarrow}$Zh} & \textbf{LoRA | Params} & \textbf{En${\rightarrow}$Zh} & \textbf{Fr${\rightarrow}$Zh}& \textbf{Hi${\rightarrow}$Zh}\\
\midrule

& & & & \textbf{Rank 8 | 20.0M} & 11.98 & 0.67 & 9.55 \\
\textbf{NeFT${_{0.4\%}}$ | 25.6M} & 8.49 & \underline{1.84} & 8.57 & \textbf{Rank 16 | 40.0M} & \textbf{\underline{13.15}} & 1.49 & \underline{9.71} \\
\hdashline[1pt/5pt]
\textbf{NeFT${_{1.5\%}}$ | 102.4M} & 2.01 & 0.91 & 9.45 & \textbf{Rank 64 | 159.9M} & \underline{11.75} & \underline{6.12} & \underline{10.01} \\
\hdashline[1pt/5pt]
\textbf{NeFT${_{3\%}}$ | 204.8M} & 1.38 & 0.39 & 9.15 & \textbf{Rank 128 | 319.8M} & \underline{11.91} & \underline{7.35} & \underline{9.18} \\
\hdashline[1pt/5pt]
\textbf{NeFT${_{6\%}}$ | 409.6M} & 2.30 & 5.18 & 9.68 & \textbf{Rank 256 | 639.6M} & \underline{10.82} & \textbf{\underline{8.10}} & 8.64 \\
\textbf{NeFT${_{9\%}}$ | 614.4M} & 1.22 & 0.47 & \underline{10.69} \\
\hdashline[1pt/5pt]
\textbf{NeFT${_{12\%}}$ | 819.2M} & 0.05 & 0.04 & \textbf{11.21} \\
\midrule
\midrule

\textbf{\# Lang En${\rightarrow}$Fr (20k)} & \textbf{BLEU} & & & & \textbf{BLEU} \\
\textbf{NeFT | Params} & \textbf{en${\rightarrow}$Fr} & \textbf{Hi${\rightarrow}$Fr} & & \textbf{LoRA | Params} & \textbf{En${\rightarrow}$Fr} & \textbf{Hi${\rightarrow}$Fr} \\
\midrule

& & & & \textbf{Rank 8 | 20.0M} & 33.13 & 1.73 \\
& & & & \textbf{Rank 16 | 40.0M} & 33.77 & 1.90 \\
& & & & \textbf{Rank 64 | 159.9M} & 33.18 & 0.26 \\
\hdashline[1pt/5pt]
\textbf{NeFT${_{3\%}}$ | 204.8M} & \underline{34.25} & \underline{3.66} & & \textbf{Rank 128 | 319.8M} & 32.98 & 0.17 \\
\hdashline[1pt/5pt]
\textbf{NeFT${_{6\%}}$ | 409.6M} & \underline{34.90} & \underline{3.84} & & \textbf{Rank 256 | 639.6M} & 33.14 & 0.07 \\
\textbf{NeFT${_{9\%}}$ | 614.4M} & 34.86 & 3.52 \\
\hdashline[1pt/5pt]
\textbf{NeFT${_{12\%}}$ | 819.2M} & \textbf{35.08} & \textbf{4.05} \\
\midrule
\midrule

\textbf{\# Lang Hi${\rightarrow}$Fr (4.5k)} & \textbf{BLEU} & & & & \textbf{BLEU} \\
\textbf{NeFT | Params} & \textbf{En${\rightarrow}$Fr} & \textbf{Hi${\rightarrow}$Fr} & & \textbf{LoRA | Params} & \textbf{En${\rightarrow}$Fr} & \textbf{Hi${\rightarrow}$Fr} \\
\midrule

& & & & \textbf{Rank 8 | 20.0M} & 20.18 & 8.00 \\
& & & & \textbf{Rank 16 | 40.0M} & \textbf{21.29} & 7.90 \\
& & & & \textbf{Rank 64 | 159.9M} & 19.22 & 7.14 \\
\hdashline[1pt/5pt]
\textbf{NeFT${_{3\%}}$ | 204.8M} & 11.40 & \underline{6.47} & & \textbf{Rank 128 | 319.8M} & \underline{18.62} & 6.05 \\
\hdashline[1pt/5pt]
\textbf{NeFT${_{6\%}}$ | 409.6M} & 12.45 & 8.59 & & \textbf{Rank 256 | 639.6M} & \underline{16.90} & 5.66 \\
\textbf{NeFT${_{9\%}}$ | 614.4M} & 9.08 & \underline{10.23} \\
\hdashline[1pt/5pt]
\textbf{NeFT${_{12\%}}$ | 819.2M} & 0.26 & \textbf{10.48} \\
\midrule
\midrule

\textbf{\# Lang En${\rightarrow}$Mi (10k)} & \textbf{BLEU} & & & & \textbf{BLEU} \\
\textbf{NeFT | Params} & \textbf{En${\rightarrow}$Mi} & & & \textbf{LoRA | Params} & \textbf{En${\rightarrow}$Mi} \\
\midrule

& & & & \textbf{Rank 8 | 20.0M} & 6.22 \\
& & & & \textbf{Rank 16 | 40.0M} & 5.88 \\
& & & & \textbf{Rank 64 | 159.9M} & 5.56 \\
\hdashline[1pt/5pt]
& & & & \textbf{Rank 128 | 319.8M} & 4.87 \\
\textbf{NeFT${_{9\%}}$ | 614.4M} & \underline{13.90} & & & \textbf{Rank 256 | 639.6M} & 4.55 \\
\hdashline[1pt/5pt]
\textbf{NeFT${_{12\%}}$ | 819.2M} & \underline{\textbf{14.04}} \\
\midrule
\midrule

\textbf{\# Lang En${\rightarrow}$Bs (10k)} & \textbf{BLEU} & & & & \textbf{BLEU} \\
\textbf{NeFT | Params} & \textbf{En${\rightarrow}$Bs} & & & \textbf{LoRA | Params} & \textbf{En${\rightarrow}$Bs} \\
\midrule

& & & & \textbf{Rank 8 | 20.0M} & 8.66 \\
& & & & \textbf{Rank 16 | 40.0M} & 8.15 \\
& & & & \textbf{Rank 128 | 319.8M} & 8.83 \\
\hdashline[1pt/5pt]
\textbf{NeFT${_{9\%}}$ | 614.4M} & \textbf{10.04} & & & \textbf{Rank 256 | 639.6M} & 8.22 \\
\bottomrule

\end{tabular}
}
\caption{\label{table10}
The BLEU scores for all LoRA settings and the comparison with NeFT of the corresponding parameter magnitude in the machine translation experiment.
}
\end{table*}
\end{document}